\documentclass[runningheads]{llncs}
\usepackage{multirow}
\usepackage{graphicx}
\usepackage{cite}
\usepackage{amsmath,amssymb,amsfonts}
\usepackage{algorithm,algorithmic}
\usepackage{graphicx}
\usepackage{textcomp}
\usepackage{subcaption}
\usepackage{graphicx}
\usepackage{multirow}
\usepackage{dtk-logos}
\usepackage{cite}
\usepackage{makecell}
\usepackage{booktabs}
\usepackage{hyperref}
\usepackage{amssymb}
\usepackage{algorithm,algorithmic}
\usepackage{lipsum}
\usepackage[english]{babel}
\usepackage[utf8]{inputenc}
\usepackage{listings}
\usepackage{graphicx}
\usepackage{amsmath}
\usepackage{cleveref}
\usepackage{enumitem}
\usepackage{hyphenat}
\usepackage{makecell}
\usepackage{adjustbox}
\usepackage{xcolor}

\begin{document}
\title{An Efficient Anomaly Detection Approach using Cube Sampling with Streaming Data}
\author{Seemandhar Jain\textsuperscript{*}\inst{1}\orcidID{0000-0002-4176-3595} \and
Prarthi Jain\textsuperscript{*}\inst{1}\orcidID{0000-0002-4947-1175} \and
Abhishek Srivastava\textsuperscript{*}\inst{1}\orcidID{0000-0002-6338-5476}
}
\authorrunning{jain et al.}
%
\institute{\textsuperscript{1}IIT Indore, India \\
\email{\{seemandharj, prarujain15\}@gmail.com, asrivastava@iiti.ac.in}\\
\textsuperscript{*} \textit{These authors contributed equally}
}%
\titlerunning{Efficient Anomaly Detection Approach using Cube Sampling}
\maketitle              
\begin{abstract}
Anomaly detection is critical in various fields, including intrusion detection, health monitoring, fault diagnosis, and sensor network event detection. The isolation forest (or \textit{iForest}) approach is a well-known technique for detecting anomalies. It is, however, ineffective when dealing with dynamic streaming data, which is becoming increasingly prevalent in a wide variety of application areas these days. In this work, we extend our previous work by proposed an efficient \textit{iForest} based approach for anomaly detection using cube sampling that is effective on streaming data. Cube sampling is used in the initial stage to choose nearly balanced samples, significantly reducing storage requirements while preserving efficiency. Following that, the streaming nature of data is addressed by a sliding window technique that generates consecutive chunks of data for systematic processing. The novelty of this paper is in applying Cube sampling in \textit{iForest} and calculating inclusion probability. The proposed approach is equally successful at detecting anomalies as existing state-of-the-art approaches, requiring significantly less storage and time complexity. We undertake empirical evaluations of the proposed approach using standard datasets and demonstrate that it outperforms traditional approaches in terms of Area Under the ROC Curve (AUC-ROC) and can handle high-dimensional streaming data.

\end{abstract}

\section{Introduction}
Anomalies are rare and distinct data patterns that vary from normal or anticipated behavior. According to Hawkins' definition \cite{hawkins1980identification}, an anomaly often referred to as an outlier is significantly different from standard observable objects and is thought to have been created by a separate mechanism. Anomaly detection seeks to identify trends that deviate from anomalies or outliers. High dimensionality complicates anomaly identification because as the number of characteristics or features increases, the amount of data required to generalize properly increases as well, resulting in data sparsity or scattered and isolated data points.
The data is primarily of the streaming variety, which refers to information that flows in and out of a device similar to a stream. Continuously developing, sequential, and unlimited streaming data consists of a perpetual flow of data pieces that evolve. This study is concerned with the identification of abnormalities in such streaming data.
Due to their large size, offline algorithms that attempt to store such streaming data are ineffective in dealing with high streams of data. The data must be handled sequentially or gradually on a data-by-data basis with the help of a sliding window technique \cite{datar2002maintaining, gibbons2002distributed}. In the case of streaming data, where data is continuously generating, the model must be continually retrained or updated utilizing the incoming continuous data stream to minimize concept drift \cite{zhou2009anomaly,gaber2005mining,gama2014survey}. Concept drift is a typical occurrence in streaming data and must be considered while doing anomaly detection. Additionally, concept drift indicates that the trained model rapidly becomes antiquated and cannot produce accurate results since it is no longer compatible with the current data.
The Isolation Forest (or \textit{iForest}) algorithm is an excellent approach for anomaly identification that manages concept drift gracefully \cite{liu2008isolation}. \textit{iForest} is a well-known randomization-based method for detecting anomalies. Randomization is a strong technique that has been demonstrated to be beneficial in supervised learning \cite{breiman2001random}. The \textit{iForest} algorithm's randomization mechanism is as follows: one attribute is picked randomly from the various features of the dataset. Following that, a random value between the range of the characteristic is selected, and the dataset is partitioned along this value. As a result of this partitioning, a binary tree is formed, with data points with values more significant than the randomly determined partitioning value constituting the right child and those with a smaller value becoming the left child. This technique is repeated until all data points are separated. Due to the rarity and uniqueness of data points that may be deemed anomalies, the likelihood of such anomalies isolating sooner and closer to the root node is greater than for normal points.
The first step in this work employs a balanced sampling approach (called cube sampling), which significantly reduces the size of the dataset and enables the iForest algorithm to be utilized effectively on streaming data with minimal computing cost. This includes the inclusion probability calculation detailed in subsequent parts. Later, we present a method that makes efficient use of the \textit{iForest} algorithm \cite{liu2008isolation} with streaming data. \textit{iForest} effectively accounts for the problem of concept drift that frequently occurs with streaming data and does not need costly distance computations, as other distances and density-based methods do. \textit{iForest} is a randomization-based algorithm that may easily be expanded to handle streaming data.
The proposed approach has several distinguishing characteristics, including the following: 1) significantly reduces the size of the data through cube sampling, allowing for detection of anomalies without the ample use of space and time; 2) effectively detects anomalies in streaming data; and 3) requires less time complexity to update the model, making it adaptable to the streaming data, thereby minimizing concept drift.

\noindent
\textbf{The following are the primary contributions of the paper.}
\begin{enumerate}
\item The paper uses the cube sampling technique to minimize the amount of the dataset while creating an \textit{iForest}.
\item The effectiveness of using a sliding window to manage flowing data is illustrated.
\item The proposed approach is evaluated on standard datasets and shown to be successful. AUC-ROC values are generally superior to and, in a few cases, equivalent to those of existing anomaly detection algorithms.
\end{enumerate}

The remainder of the paper is structured in the following manner. Section \ref{sec:rw} refers to related work and other significant contributions to the field of anomaly detection. The suggested methodology is detailed in Section \ref{sec:pa}, which contains a description of the cube sampling phase, the anomaly detection step, and the update step.
Experimental analysis is performed in Section \ref{sec:analysis} on a variety of real-world datasets as well as on a synthetically generated dataset. Finally, Section \ref{sec:concl} summarizes the paper and makes recommendations for further work.

 \begin{figure}
\centering 
\captionsetup{justification=centering}
\includegraphics[width=0.8\linewidth]{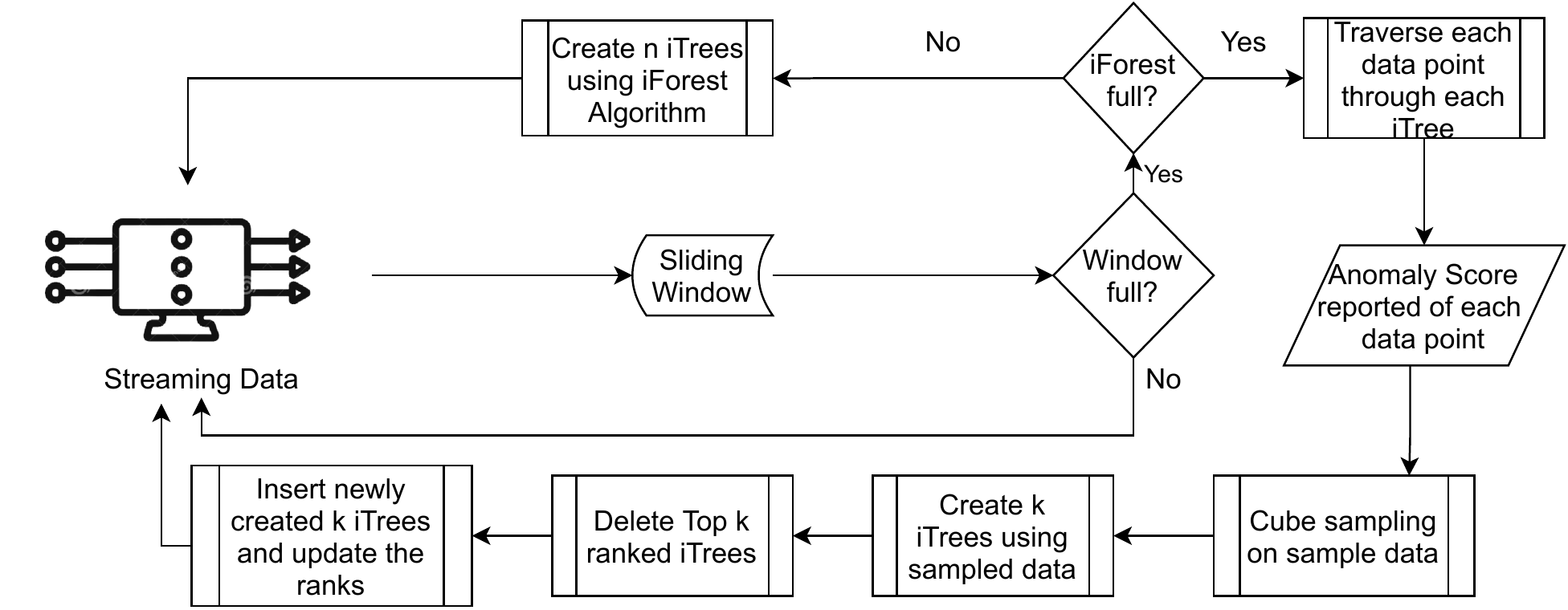}
\caption{Framework of the proposed approach for Streaming Data}
\label{fig:framework2}
\end{figure}

\section{Related Work}\label{sec:rw}
Anomaly detection, also known as outlier identification, is the study of data patterns that do not comply with typical or expected behaviour. A comprehensive examination and survey of anomaly detection are accessible in \cite{hawkins1980identification, chandola2009anomaly,chandola2008comparative}. A substantial amount of research has been conducted on the use of tree-based algorithms for outlier identification; notable examples are \textit{iForest} \cite{liu2008isolation} and Extended \textit{iForest} \cite{xu2017improved, liu2018optimized}. While these efforts are beneficial for static data, they are ineffective for streaming data.
To deal with constant data generation, model updates are required to avoid concept drift. We selected the balanced sample and updated the model using cube sampling. Sampling is the process of choosing a subset of the population to reflect the entire population accurately. Sampling techniques are classified as Probability Sampling techniques \cite{cochran2007sampling} and Non-Probability Sampling techniques \cite{vehovar2016non}. \cite{taherdoost2016sampling} for a thorough study of sampling approaches. Prior to using unequal probability sampling, an inclusion probability must be determined. There have been efforts to assess the probability of inclusion for large datasets, such as Jain et al. \cite{jain2021cube} and Nigam et al. \cite{nigam1984some}. This is not a feasible strategy for dealing with multi-dimensional data.
This paper presents an improvised use of the \textit{iForest} Algorithm based on cube sampling that is both effective and memory-efficient for detecting anomalies in streaming data.

\section{Proposed Approach}\label{sec:pa}
 In this paper, we extend our previous work \cite{jain2021anomaly} and used the well-known \textit{iForest} anomaly detection technique on streaming data, followed by the cube sampling approach to select the balanced sample to update the model, which makes the approach effective for high dimensional data and mitigates concept drift. 
 
 \begin{algorithm}[H]

 \caption{$Anomlay Detection$($ntrees$, $\omega = 256$, $X$,$ktrees=10$)}
 \begin{algorithmic}[1]
  \label{al2}
 \renewcommand{\algorithmicrequire}{\textbf{Input:}}
 \renewcommand{\algorithmicensure}{\textbf{Output:}}
 \REQUIRE $X (Streaming Data) =\left\{X_{1}, X_{2}, \ldots, X_{i}, \ldots,\right\},$  $\omega-$ Sliding Window Size (=256), $ntrres$ - No. of iTree in forest(=50), $ktrees$ - No. of iTree to be inserted and deleted, to update the model 
 \ENSURE  Anomalies
 \\
     \STATE $\textit{Initialize Forest  $F$ } \gets \textit{$[]$}$
    \STATE $\textit{Sliding Window $Y$} \gets \textit{$[]$}$
    \STATE $\textit{Initialize $c$=$0$, which counts the number of data-points}$
    \STATE $\textit{while $c$ != $\omega$  do}$ \par
        \hskip\algorithmicindent $\textit{$c$} \gets \textit{$c+1$}$ \par
        \hskip\algorithmicindent $\textit{Add $X_{i}$ in $Y$ }$ \par
    \STATE $\textit{end while}$ 
    \STATE $\textit{Initialize the height of iTree $h$} \gets \textit{$ceil\left( log_{2}(\omega) \right)$}$
    \STATE $\textit{$\pi$ = Inclusion Probability proportional to t-SNE\_reduction\_technique(Y)}$ 
    \STATE $\textit{S} \gets  \textit{Cube Sampling($Y, \pi$)}$ \par
	\STATE $\textit{for $i$} \gets \textit{$0$ to $ntrees$ do}$ \par
        \hskip\algorithmicindent $\textit{$F[i]$} \gets \textit{$iTree$($S$,$0$,$h$)}$ \par 
    \STATE $\textit{end for}$
 
    \STATE $\textit{Re-initialize $c$} \gets \textit{$\omega$} \text{ $and$ } \textit{$Y$} \gets \textit{$[]$  }$
    \STATE $\textit{while $c$ $>$ $0$  do}$ \par
    
    \hskip\algorithmicindent $\textit{$c$} \gets \textit{$c-1$}$ \par 
    \hskip\algorithmicindent $\textit{Add $X_{i}$ in $Y$}$ \par 
    \STATE $\textit{end while}$ 
    \STATE $\textit{\text{report anomaly detector} $G($Y$)$ }$ \par   
    \STATE $\textit{$\pi$ = Inclusion Probability proportional to t-SNE\_reduction\_technique(Y)}$ 
     \STATE $\textit{S} \gets  \textit{Cube Sampling($Y, \pi$)}$ \par
	  \STATE $\textit{for $i$} \gets \textit{top rank trees do}$ \par
	  
        \hskip\algorithmicindent $\textit{delete $F[\textit{$i$}]$}$ \par
        \hskip\algorithmicindent $\textit{$F[\textit{$i$}]$} \gets \textit{$iTree$($S$,$0$,$h$)}$ \par 
\hskip\algorithmicindent $\textit{Assign ranks to newly created iTrees}$ \par
    
    \STATE $\textit{end for}$
	
 \STATE $\textit{goto 10 for upcoming Streaming data points}$ 
 \end{algorithmic}
 \end{algorithm}

 Due to the continuous nature of streaming data, any offline approach would run out of memory if it tries to save the whole data. Because streaming data changes over time, the model must be updated on a frequent basis to minimize concept drift (as previously stated) and to maintain a high degree of accuracy.

The rank of each element in the forest is directly proportional to the number of anomalous points detected. So lower the number of anomalies, the better the $iTree$ performance. Hence, when the model is updated, the high-rank trees are deleted first. The anomaly detection in streaming data identifies points that are out of the typical or abnormal. The figure \ref{fig:framework2} illustrates the basic framework for detecting anomalies in streams. The streaming data is initially sent through a sliding window. The complete process of anomaly detection in streaming data is explained in detail in the Algorithm  \ref{al2}. A sliding window is a helpful technique for doing calculations on streaming data in such a way that only blocks of data containing $\omega$ items of the stream are evaluated at a time.

\section{Experimental Results}\label{sec:analysis}

The proposed methodology to detect anomalies is evaluated in the following manner: 1) The proposed approach is compared with other state-of-art anomaly detection approaches in terms of \textit{AUC-ROC}; 2) Finally, the proposed approach is evaluated on a variety of anomaly patterns artificially introduced into the data stream.

In our experiments, we utilize the formula in Equation \ref{form:auc} (from \cite{liu2008isolation}) to compute the value of \textit{AUC-ROC}.

\begin{equation}\label{form:auc} 
   \textit{AUC-ROC}=\left(\Sigma r_{i}-\left(n_{\alpha}^{2}+n_{a}\right) / 2\right) /\left(n_{\alpha}\times n_{n}\right)
\end{equation}
\noindent
Where $n_{a}$ represents the number of actual anomalies, $n_{n}$ denotes the number of actual normal points, and $r_{i}$ denotes the rank of the $i^{th}$ anomaly in the descending ranked list of anomalies. There are other metrics such as Kappa statistics, which can be utilized to calculate the efficacy of the proposed approach, but it has its own limitations \cite{powers2012problem}.

\begin{table}
\centering
\caption{Dataset information}
\label{tab:dataset2}
\begin{tabular}{|l|c|c|c|} 
\hline
Dataset        & no. of records & no. of attributes & Anomaly threshold  \\ 
\hline
Mulcross       & 262144         & 4                 & 10.00\%            \\
Forest Cover   & 286048         & 10                & 0.90\%             \\
Breastw        & 683            & 9                 & 35.00\%            \\
Http(KDDcup99) & 567497         & 3                 & 0.39\%             \\
Satellite      & 6435           & 36                & 32.00\%            \\
Shuttle        & 49097          & 9                 & 7.15\%             \\
\hline
\end{tabular}
\end{table}

Five real-world datasets and one artificial dataset were used in the experiments \cite{frank2010uci}. Table \ref{tab:dataset2} contains a brief summary of the datasets utilized. The table's anomaly threshold column indicates the proportion of abnormalities in the dataset. The tests were conducted on these datasets because they contain pre-defined anomalies that serve as class labels, making them suitable for assessment. This is precisely why these datasets are often regarded as industry standards for anomaly detection.

The experiments were conducted on a personal computer equipped with an Intel(R) Core(TM) \textit{i7-7500U} processor running at $@2.70GHz$, $2.90GHz$, and $16.0GB$ of memory (RAM), on Windows $10$ Pro using Python $3.6$.

\subsection{Comparison of the Proposed Approach with state-of-art approaches}\label{sec:compare}

The proposed approach is compared with a few state-of-art anomaly detection approaches for streaming data: 1) iForest\cite{liu2008isolation}, 2) RRCF\cite{guha2016robust},  3) HSTa\cite{tan2011fast}, and 4) PiForest\cite{jain2021anomaly}.
Table \ref{spacetable} shows the comparison of the proposed approach with other anomaly detection techniques in terms of space complexities. The comparison in terms of AUC is shown in Table \ref{spacetable1}. The proposed approach works well on streaming data, and the results show that the proposed approach out-performs state-of-art algorithms for streaming data.

\begin{table}
\centering
\caption{Comparison of space complexity of state-of-art algorithms}
\label{spacetable}\resizebox{\linewidth}{!}{%
\begin{tabular}{|c|c|c|c|c|c|} 
\hline
                                                                                      & Proposed Approach                                                                                    & PiForest                                                                                             & iForest                                                                                              & RRCF                                                                                              & HSTa                                                                        \\ 
\hline
Space complexity                                                                      & O($\Psi$tb)                                                                                          & O($\Psi$tb)                                                                                          & O($\Psi$tb)                                                                                          & O($\Psi$tb)                                                                                       & O(t2\textsuperscript{h})                                                    \\ 
\hline
Parameters                                                                            & \begin{tabular}[c]{@{}c@{}}$\Psi$-Sub-sampling size\\ t-No. of trees\\ b-Size of a node\end{tabular} & \begin{tabular}[c]{@{}c@{}}$\Psi$-Sub-sampling size\\ t-No. of trees\\ b-Size of a node\end{tabular} & \begin{tabular}[c]{@{}c@{}}$\Psi$-Sub-sampling size\\ t-No. of trees\\ b-Size of a node\end{tabular} & \begin{tabular}[c]{@{}c@{}}$\Psi$- Sample size\\ t- No. of trees\\ b- Size of a node\end{tabular} & \begin{tabular}[c]{@{}c@{}}t- No. of trees\\ h- Depth of tree\end{tabular}  \\ 
\hline
\begin{tabular}[c]{@{}c@{}}Parameter values(Default\\ in the experiment)\end{tabular} & \begin{tabular}[c]{@{}c@{}}$\Psi$=256\\ t=50\\ b depends on data\end{tabular}                        & \begin{tabular}[c]{@{}c@{}}$\Psi$=511\\ t=10\\ b=4.125 Bytes\end{tabular}                            & \begin{tabular}[c]{@{}c@{}}$\Psi$=511\\ t=100\\ b depends on data\end{tabular}                       & \begin{tabular}[c]{@{}c@{}}$\Psi$=1000\\ t=200\\ b depends on data\end{tabular}                   & \begin{tabular}[c]{@{}c@{}}t=25\\ h=15\end{tabular}                         \\
\hline
\end{tabular}}
\end{table}

\subsection{The proposed approach's performance on a variety of anomaly types}\label{sec:anomalypatterns}
\begin{figure} 
\centering 
\captionsetup{justification=centering}
\includegraphics[width=0.8\linewidth]{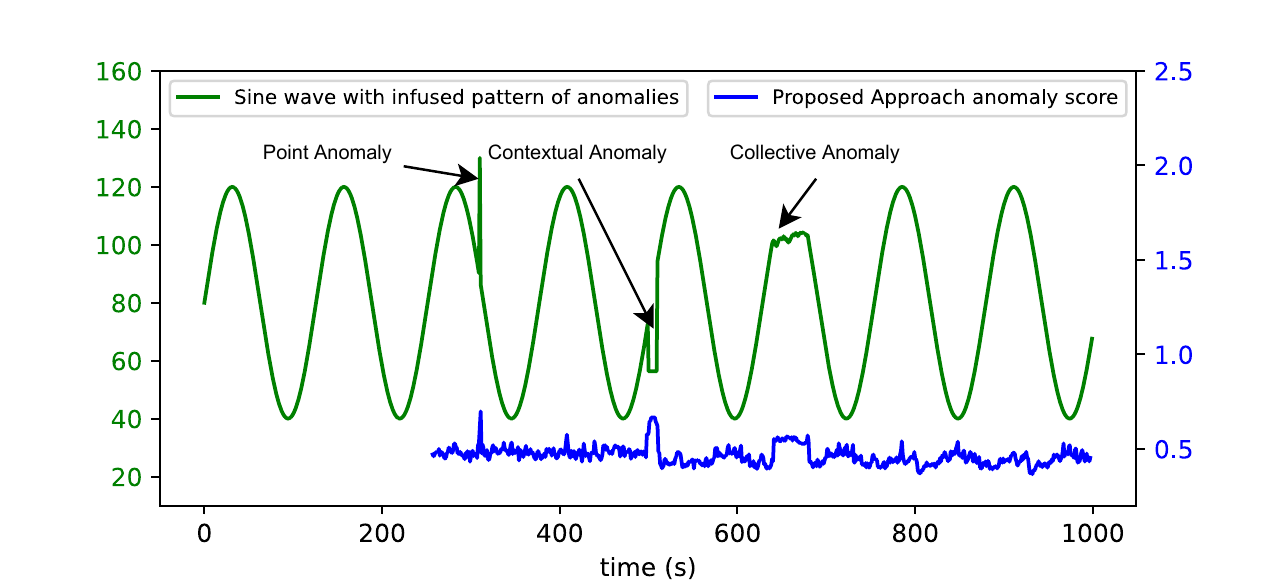}
\caption{Anomaly Score with generated data}
\label{fig:setup2}
\end{figure}

We evaluate the proposed approach's effectiveness against a variety of patterns of anomalies \cite{chandola2009anomaly} that occur often in streams. We produced a synthetic dataset by injecting anomalies into a sine wave. There are three types of anomalies: point anomaly, contextual anomaly , and collective anomaly . These anomalies are injected as shown in Figure \ref{fig:setup2}. We manage the streaming data using a method termed shingling, similar to Guha \textit{et al.} \cite{guha2016robust}. A shingle size of $n$ placed across a stream collects the stream's initial $n$ data points from time $T=0, T=1,...,T=N$ to generate a $N$-dimensional data point. Following that, at time $T=(N+1)$, the data from $T=2,..., t=(N+1)$ are recorded to construct a second $N$-dimensional data point. A shingle gives the curve a characteristic form, and any variation from that shape signals an abnormality.

\begin{table}[]
\centering
\caption{Comparison of the proposed approach with state-of-art algorithms in terms of AUC}
\label{spacetable1}
\begin{tabular}{|l|c|c|c|l|l|}
\hline
Dataset        & \multicolumn{1}{l|}{Proposed Approach} & \multicolumn{1}{l|}{iForest} & \multicolumn{1}{l|}{PiForest} & HSTa & RRCF \\ \hline
Http(KDDcup99) & 0.99                                   & 0.99                         & 0.99                          & 0.96 & 0.97 \\
Satellite      & 0.72                                   & 0.70                         & 0.67                          & 0.66 & 0.64 \\
Mulcross       & 0.98                                   & 0.97                         & 0.87                          & 0.86 & 0.87 \\
Forest Cover   & 0.93                                   & 0.91                         & 0.70                          & 0.74 & 0.65 \\
Breastw        & 0.96                                   & 0.95                         & 0.88                          & 0.81 & 0.89 \\
Shuttle        & 0.99                                   & 0.99                         & 0.99                          & 0.98 & 0.99 \\ \hline
\end{tabular}
\end{table}

In Figure \ref{fig:setup2}, the curve below the sine wave shows the anomaly score obtained using the proposed method.
The first $256$ samples are utilized for training the anomaly detector, and the succeeding data stream is used to compute the anomaly scores. The following classification accuracy metrics are collected for specific anomalies: Contextual Anomaly (0.99); Point Anomaly (1.00); and Collective Anomaly (0.962). Moreover, an \textit{AUC-ROC} score of $0.99$ is obtained, indicating that the suggested method is effective with a variety of anomaly patterns. When tested on \textit{iForest} trivially, an \textit{AUC-ROC} score of $0.83$ is recorded, suggesting that \textit{iForest} is not effective in handling streaming data.

\section{Conclusion}\label{sec:concl}
In this paper, we present a method for detecting anomalies utilizing the \textit{iForest} algorithm and cube sampling. Additionally, the method permits the processing of streaming data, which is data that flows constantly and may be regarded as limitless in all practical senses. We handle such data well by utilizing a sliding window. The approach's efficacy in terms of anomaly identification is demonstrated by comparison to the operation of various well-known anomaly detection algorithms. In each example, despite dealing with limited data, the strategy produces findings that are equal to or superior to those obtained using existing anomaly detection approaches. Additionally, we show the proposed approach works well in a simulated environment.

\bibliographystyle{splncs04}
\bibliography{Ref}

\begin{thebibliography}{10}
\providecommand{\url}[1]{\texttt{#1}}
\providecommand{\urlprefix}{URL }
\providecommand{\doi}[1]{https://doi.org/#1}

\bibitem{breiman2001random}
Breiman, L.: Random forests. Machine learning  \textbf{45}(1),  5--32 (2001)

\bibitem{chandola2009anomaly}
Chandola, V., Banerjee, A., Kumar, V.: Anomaly detection: A survey. ACM
  computing surveys (CSUR)  \textbf{41}(3),  1--58 (2009)

\bibitem{chandola2008comparative}
Chandola, V., Mithal, V., Kumar, V.: Comparative evaluation of anomaly
  detection techniques for sequence data. In: 2008 Eighth IEEE international
  conference on data mining. pp. 743--748. IEEE (2008)

\bibitem{cochran2007sampling}
Cochran, W.G.: Sampling techniques. John Wiley \& Sons (2007)

\bibitem{datar2002maintaining}
Datar, M., Gionis, A., Indyk, P., Motwani, R.: Maintaining stream statistics
  over sliding windows. SIAM journal on computing  \textbf{31}(6),  1794--1813
  (2002)

\bibitem{frank2010uci}
Frank, A.: Uci machine learning repository. http://archive. ics. uci. edu/ml
  (2010)

\bibitem{gaber2005mining}
Gaber, M.M., Zaslavsky, A., Krishnaswamy, S.: Mining data streams: a review.
  ACM Sigmod Record  \textbf{34}(2),  18--26 (2005)

\bibitem{gama2014survey}
Gama, J., {\v{Z}}liobait{\.e}, I., Bifet, A., Pechenizkiy, M., Bouchachia, A.:
  A survey on concept drift adaptation. ACM computing surveys (CSUR)
  \textbf{46}(4),  1--37 (2014)

\bibitem{gibbons2002distributed}
Gibbons, P.B., Tirthapura, S.: Distributed streams algorithms for sliding
  windows. In: Proceedings of the fourteenth annual ACM symposium on Parallel
  algorithms and architectures. pp. 63--72. ACM (2002)

\bibitem{guha2016robust}
Guha, S., Mishra, N., Roy, G., Schrijvers, O.: Robust random cut forest based
  anomaly detection on streams. In: International conference on machine
  learning. pp. 2712--2721 (2016)

\bibitem{hawkins1980identification}
Hawkins, D.M.: Identification of outliers, vol.~11. Springer (1980)

\bibitem{jain2021anomaly}
Jain, P., Jain, S., Za{\"\i}ane, O.R., Srivastava, A.: Anomaly detection in
  resource constrained environments with streaming data. IEEE Transactions on
  Emerging Topics in Computational Intelligence  (2021)

\bibitem{jain2021cube}
Jain, S., Shastri, A.A., Ahuja, K., Busnel, Y., Singh, N.P.: Cube sampled
  k-prototype clustering for featured data. arXiv preprint arXiv:2108.10262
  (2021)

\bibitem{liu2008isolation}
Liu, F.T., Ting, K.M., Zhou, Z.H.: Isolation forest. In: 2008 Eighth IEEE
  International Conference on Data Mining. pp. 413--422. IEEE (2008)

\bibitem{liu2018optimized}
Liu, Z., Liu, X., Ma, J., Gao, H.: An optimized computational framework for
  isolation forest. Mathematical Problems in Engineering  \textbf{2018},  1--13
  (2018)

\bibitem{nigam1984some}
Nigam, A., Kumar, P., Gupta, V.: Some methods of inclusion probability
  proportional to size sampling. Journal of the Royal Statistical Society:
  Series B (Methodological)  \textbf{46}(3),  564--571 (1984)

\bibitem{powers2012problem}
Powers, D.M.W.: The problem with kappa. In: Proceedings of the 13th Conference
  of the European Chapter of the Association for Computational Linguistics. pp.
  345--355 (2012)

\bibitem{taherdoost2016sampling}
Taherdoost, H.: Sampling methods in research methodology; how to choose a
  sampling technique for research. How to Choose a Sampling Technique for
  Research (April 10, 2016)  (2016)

\bibitem{tan2011fast}
Tan, S.C., Ting, K.M., Liu, T.F.: Fast anomaly detection for streaming data.
  In: Twenty-Second International Joint Conference on Artificial Intelligence.
  pp. 1511--1516 (2011)

\bibitem{vehovar2016non}
Vehovar, V., Toepoel, V., Steinmetz, S.: Non-probability sampling. The Sage
  handbook of survey methods pp. 329--345 (2016)

\bibitem{xu2017improved}
Xu, D., Wang, Y., Meng, Y., Zhang, Z.: An improved data anomaly detection
  method based on isolation forest. In: 2017 10th International Symposium on
  Computational Intelligence and Design (ISCID). vol.~2, pp. 287--291. IEEE
  (2017)

\bibitem{zhou2009anomaly}
Zhou, J., Fu, Y., Wu, Y., Xia, H., Fang, Y., Lu, H.: Anomaly detection over
  concept drifting data streams. Journal of Computational Information Systems
  \textbf{5}(6) (2009)

\end{thebibliography}
\end{document}